\documentclass[suppldata]{interact}
\usepackage{subfigure}% Support for small, `sub' figures and tables
\usepackage{natbib}% Citation support using natbib.sty
\bibpunct[, ]{(}{)}{;}{a}{}{,}% Citation support using natbib.sty
\renewcommand\bibfont{\fontsize{10}{12}\selectfont}% Bibliography support using natbib.sty
\theoremstyle{remark}

\usepackage{hyperref,xcolor}
\usepackage{multirow}
\usepackage{cite}
\usepackage{lscape}
\usepackage{afterpage}
\usepackage{graphicx}
\usepackage{academicons}
\usepackage{xcolor}
\theoremstyle{plain}% Theorem-like structures provided by amsthm.sty

\theoremstyle{definition}

\begin{document}
%\articletype{ARTICLE TEMPLATE}
\begin{titlepage}
\title{Log Message Anomaly Detection and Classification Using Auto-B/LSTM and Auto-GRU}
\author{
\name{Amir Farzad\textsuperscript{a,*,1}\thanks{\textsuperscript{*,1} Amir Farzad. Email: amirfarzad@uvic.ca. ORCiD: 0000-0003-2499-7696} and T. Aaron Gulliver\textsuperscript{a,2}\thanks{\textsuperscript{2}T. Aaron Gulliver. Email: agullive@ece.uvic.ca. ORCiD: 0000-0001-9919-0323}}
\affil{\textsuperscript{a}Department of Electrical and Computer Engineering, University of Victoria, PO Box 1700, STN CSC, Victoria, BC Canada V8W 2Y2}
}
\maketitle
\end{titlepage}
\clearpage
\title{Log Message Anomaly Detection and Classification Using Auto-B/LSTM and Auto-GRU}
\maketitle
\begin{abstract}
Log messages are now widely used in software systems.
They are important for classification as millions of logs are generated each day.
Most logs are unstructured which makes classification a challenge.
In this paper, Deep Learning (DL) methods called Auto-LSTM, Auto-BLSTM and Auto-GRU are developed for anomaly detection and log classification.
These models are used to convert unstructured log data to extracted features which is suitable for classification algorithms.
They are evaluated using four data sets, namely BGL, Openstack, Thunderbird and IMDB.
The first three are popular log data sets while the fourth is a movie review data set which is used for sentiment classification.
The results obtained show that Auto-LSTM, Auto-BLSTM and Auto-GRU perform better than other well-known algorithms.
\end{abstract}


\section{Introduction} \label{section.intro}
Software frameworks and systems such as web search engines and cloud computing servers are now prevalent in everyday life.
Availability is expected 24/7 and access failures can cause significant hardship to both organizations and clients.
In a framework, each logging statement produces log messages corresponding to a particular task.
Unstructured log messages contain runtime data including verbosity level, timestamp (event time), and raw message content which is a text description of the framework activity.
The structure of these messages can vary significantly between frameworks which makes anomaly detection in unstructured logs difficult \citep{yuanSherLogErrorDiagnosis2010}.

Logs which record runtime data are used for many tasks including anomaly detection \citep{linLogClusteringBased2016}, program verification \citep{dingLog2CostAwareLogging2015}, and performance monitoring \citep{nagarajStructuredComparativeAnalysis2012}.
Some techniques use rules for detecting anomalies in log messages but these require explicit domain knowledge \citep{yenBeehiveLargescaleLog2013}.
Most consider just a single component of a log message such as the verbosity level which restricts the anomalies that can be identified
\citep{yuCloudSeerWorkflowMonitoring2016}.
These tasks can be done manually on a small scale, but such anomaly detection depends on extensive manual review of logs and a keyword search
cannot be used for detecting suspicious log messages for large scale frameworks \citep{linLogClusteringBased2016}.
Modern frameworks produce huge amounts of log data, e.g. at a rate of over 50 Gigabytes (millions of lines) per hour \citep{miFineGrainedUnsupervisedScalable2013}.
Thus, automated log analysis strategies for anomaly detection and classification are necessary.

An Autoencoder \citep{rumelhartParallelDistributedProcessing1986} is a feed forward Artificial Neural Network (ANN) that can can learn features from data and the data structure \citep{vincentExtractingComposingRobust2008,rifaiContractiveAutoencodersExplicit2011}.
Autoencoders have been applied to many different tasks such as probabilistic and generative modeling \citep{kingmaAutoEncodingVariationalBayes2013,rezendeStochasticBackpropagationApproximate2014}, representation learning \citep{higginsBetaVAELearningBasic2017},
and interpolation \citep{meschederAdversarialVariationalBayes2017a,robertsHierarchicalLatentVector2018a}.
A Long Short-Term Memory (LSTM) network \citep{hochreiterLongShortTermMemory1997} is a Recurrent Neural Network (RNN)
that uses a cell to retain sequence information and remember long-term dependencies.
LSTMs have been successfully used for tasks such as language modeling \citep{zarembaRecurrentNeuralNetwork2014},
translation \citep{luongAddressingRareWord2015},
analysis of audio and video data \citep{marchiMultiresolutionLinearPrediction2014,donahueLongTermRecurrentConvolutional2017},
phoneme classification \citep{gravesBidirectionalLSTMNetworks2005},
online mode-detection \citep{otteLocalFeatureBased2012},
emotion recognition \citep{wollmerContextsensitiveMultimodalEmotion2010}
and acoustic modeling \citep{sakLongShorttermMemory2014}.
An LSTM only utilizes past context but a Bidirectional Long-Short Term Memory (BLSTM) \citep{gravesFramewisePhonemeClassification2005}
can utilize both the past and future contexts by processing the input data in both the forward and backward directions.
BLSTMs have have been used for numerous tasks such as text recognition and classification \citep{chavanPrintedTextRecognition2017,zhouTextClassificationImproved2016}.
A Gated Recurrent Unit (GRU) \citep{choLearningPhraseRepresentations2014} is similar to an LSTM network and the performance is comparable.
However, it has been shown to provide better results than an LSTM in tasks such as speech recognition \citep{irieLSTMGRUHighway2016}.

Log messages can be considered as sequences so an LSTM or GRU network with an Autoencoder is a suitable structure.
Thus in this paper, models are proposed called Auto-LSTM, Auto-BLSTM and Auto-GRU that first extract features from log messages using an Autoencoder
and then use the resulting extracted features in an LSTM, BLSTM or GRU network for anomaly detection and classification.
The models are evaluated based on the accuracy, precision, recall and F-measure using
three labeled log message data sets, namely BlueGene/L (BGL)\footnote{\url{https://github.com/logpai/loghub/tree/master/BGL}}, Openstack\footnote{\url{https://github.com/logpai/loghub/tree/master/OpenStack}}
and Thunderbird\footnote{\url{https://github.com/logpai/loghub/tree/master/Thunderbird}}.
The IMDB movie review data set\footnote{\url{https://ai.stanford.edu/~amaas/data/sentiment}}
is also considered for sentiment classification.
It is shown that good results are obtained with the proposed models for all four data sets with the same configuration.

The main contributions are as follows.
\begin{enumerate}
\item The Auto-LSTM, Auto-BLSTM and Auto-GRU models are proposed for anomaly detection and classification.
\item The proposed models are evaluated using four data sets and the results are compared with those for other well-known models.
This shows that the proposed models provide the best performance.
\end{enumerate}
The rest of the paper is organized as follows.
Section \ref{section.model} presents the Autoencoder, LSTM, BLSTM and GRU architectures followed by the proposed models.
The simulation results and discussion are given in Section \ref{section.results}.
Finally, Section \ref{section.conclusion} provides some concluding remarks.

\section{System Model} \label{section.model}

In this section, the Autoencoder, LSTM, BLSTM and GRU architectures employed in the proposed models are described.

\subsection{Autoencoder Architecture} \label{section.Autoencoder}

\begin{figure}
\centering
\includegraphics[scale=0.5]{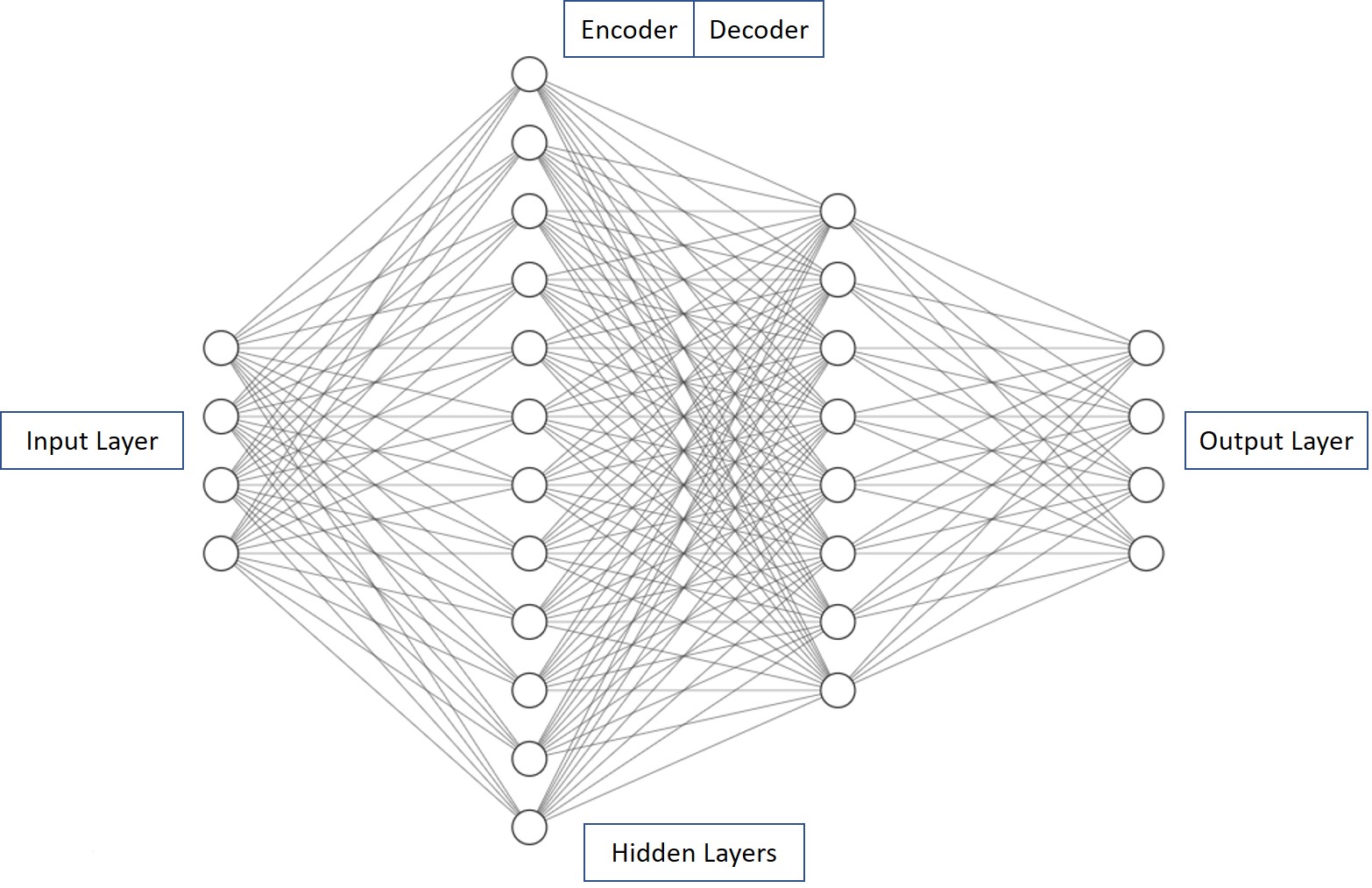}
\caption{Autoencoder structure with an input layer, output layer and two hidden layers.} \label{fig.1}
\end{figure}

An Autoencoder \citep{rumelhartParallelDistributedProcessing1986} is a feed-forward multilayer neural network with the same number of input and output neurons.
The goal is to learn a compressed representation while minimizing the error for the input data.
Training is done using the backpropagation algorithm according to a loss function.
An Autoencoder with more than one hidden layer is called a deep Autoencoder \citep{lecunDeepLearning2015}.
Having many encoder and decoder layers enables a deep Autoencoder to represent complex input distributions.
Figure~\ref{fig.1} shows the architecture of an Autoencoder with an input layer, an output layer, and two hidden layers.

\subsection{LSTM Architecture} \label{section.LSTM}

\begin{figure}
\centering
\includegraphics[scale=0.5]{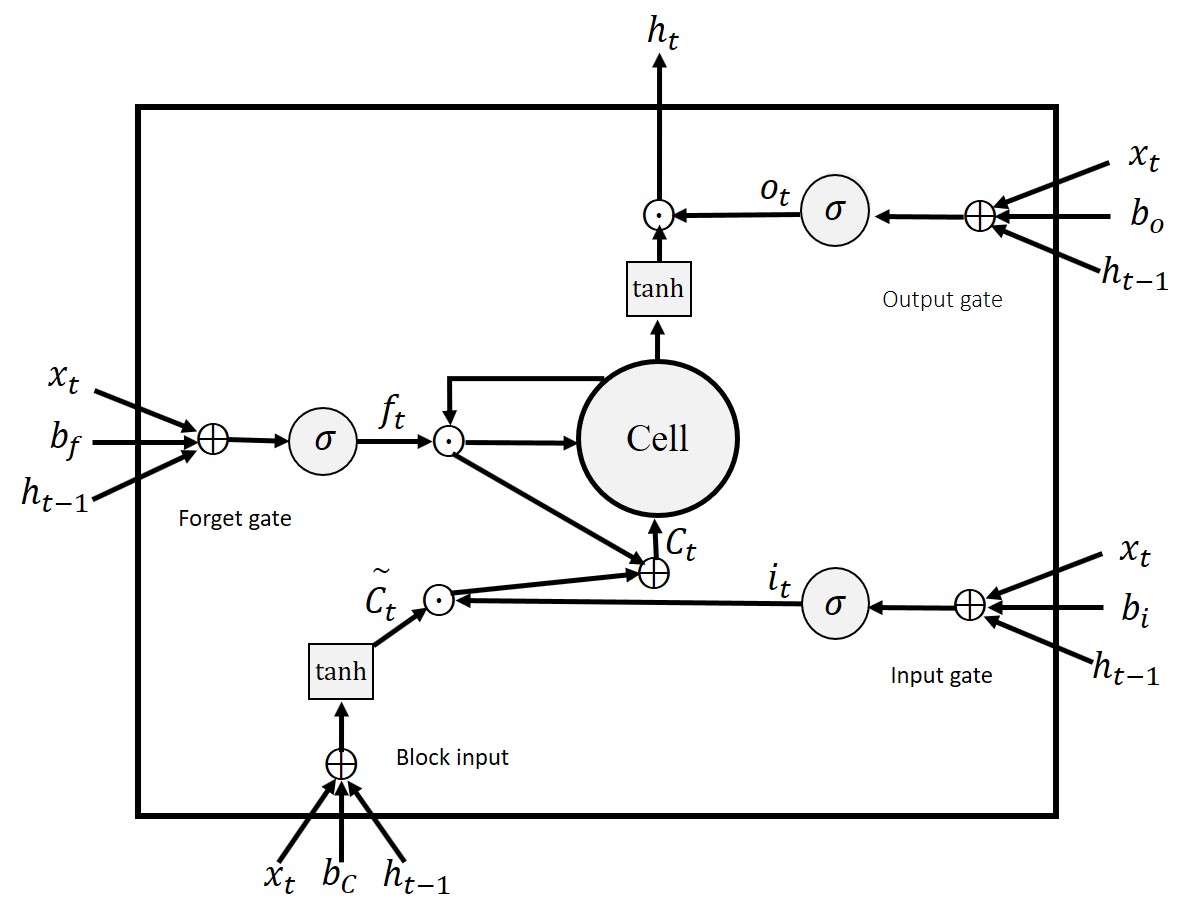}
\caption{A LSTM block with input gate, output gate, forget gate, block input, and sigmoid and hyperbolic tangent activation functions.} \label{fig.2}
\end{figure}

An LSTM is a recurrent neural network (RNN) \citep{hochreiterLongShortTermMemory1997} which has been successfully used to solve sequential data problems \citep{gravesSupervisedSequenceLabelling2012}.
It has cells to store information in blocks which can be recurrently connected.
These cells solve the vanishing gradient problem.
Each LSTM block contains self-connected cells with input, forget, and output gates.
These gates are designed to store information longer than feedforward neural networks to improve performance \citep{gravesSupervisedSequenceLabelling2012}.
A block of an LSTM network contains recurrently connected cells as shown in Figure~\ref{fig.2}.
The cell input at time $t$ is $x_t$ and the corresponding values for
the input, forget and output gates are
\begin{equation} \label{eq.1}
i_t=\sigma(W_ix_t+U_ih_{t-1}+b_i),
\end{equation}
\begin{equation} \label{eq.2}
f_t=\sigma(W_fx_t+U_fh_{t-1}+b_f),
\end{equation}
\begin{equation} \label{eq.3}
o_t=\sigma(W_ox_t+U_oh_{t-1}+b_o),
\end{equation}
respectively, where $W$ and $U$ are the weight matrices, $b$ is the bias vector, and
$\sigma$ is the sigmoid activation function.
The block input at time $t$ is
\begin{equation} \label{eq.4}
\hat{C_t}=\tanh(W_Cx_t+U_Ch_{t-1}+b_C),
\end{equation}
where $W_C$ and $U_C$ are the weight matrices, $b_C$ is the bias vector, and
tanh denotes the hyperbolic tangent activation function.
The cell state  at time $t$ is
\begin{equation} \label{eq.5}
C_t=f_t \odot C_{t-1}+i_t \odot \hat{C_t},
\end{equation}
where $\odot$ denotes point-wise multiplication.
Finally, the block output at time $t$ is given by
\begin{equation} \label{eq.6}
h_t=o_t \odot \tanh(C_t).
\end{equation}

\subsection{BLSTM Architecture} \label{section.BLSTM}

\begin{figure}
\centering
\includegraphics[scale=0.5]{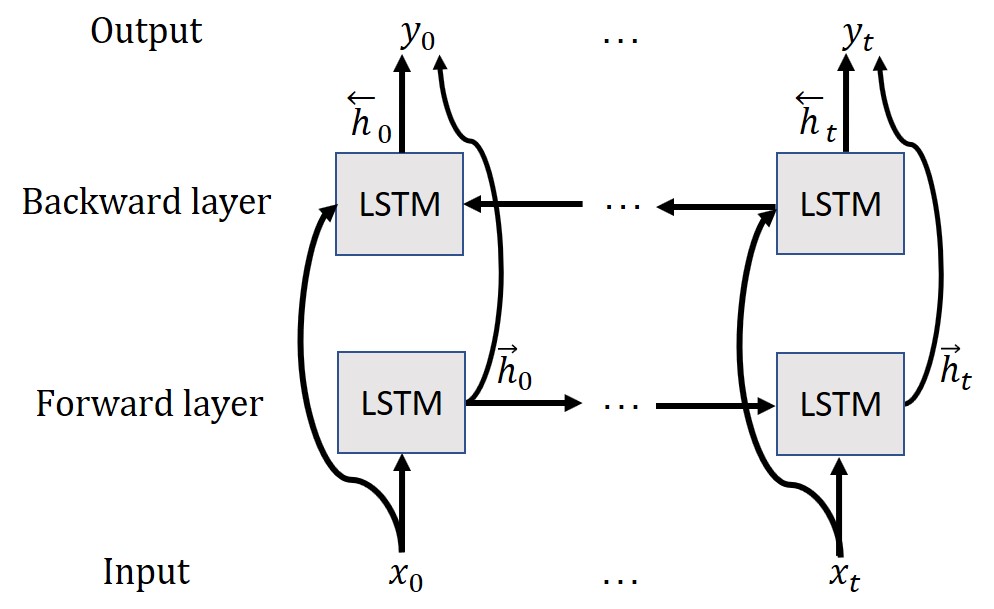}
\caption{BLSTM architecture with forward and backward LSTM layers.} \label{fig.3}
\end{figure}

Bidirectional Long-Short Term Memory (BLSTM) consists of two LSTM networks.
The input data is fed into two LSTM networks forwards and backwards with respect to time $t$, respectively, and both are connected to the same output layer.
BLSTMs have the benefit of using both the past context and future context in a sequence.
In a BLSTM, the forward and backward LSTM outputs at time $t$ are concatenated which is expressed by
\begin{equation} \label{eq.7}
h_t=[\overrightarrow{h_t},\overleftarrow{h_t}],
\end{equation}
where $\overrightarrow{h_t}$ is the forward block output and $\overleftarrow{h_t}$ is the backward block output.
The final output at time $t$ is
\begin{equation} \label{eq.8}
y_t=\sigma(W_yh_t+b_y),
\end{equation}
where $W_y$ is the weight matrix, $b_y$ is the bias vector and $\sigma$ is the activation function.
The BLSTM network architecture with forward and backward LSTM layers is shown in Figure~\ref{fig.3}.

\subsection{GRU Architecture} \label{section.GRU}

\begin{figure}
\centering
\includegraphics[scale=0.5]{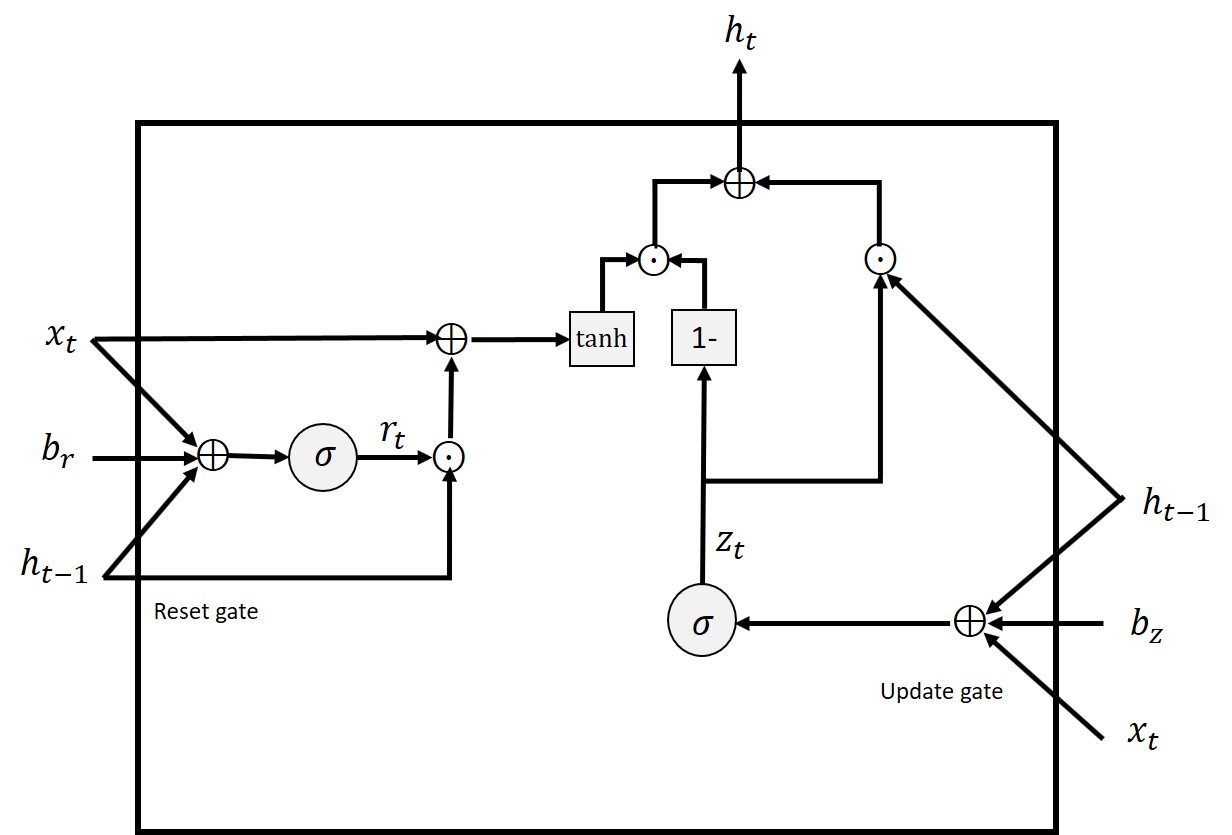}
\caption{A GRU block with reset gate, update gate, sigmoid and hyperbolic tangent activation functions.} \label{fig.4}
\end{figure}

A Gated Recurrent Unit (GRU) is a modified LSTM network.
The difference is that a GRU has a reset gate and an update gate.
A block of a GRU network is shown in Figure~\ref{fig.4}.
In this block, the reset gate is used to decide how much is forgotten.
The reset gate is expressed by
\begin{equation} \label{eq.9}
r_t = \sigma(W_rx_t+U_rh_{t-1}+b_r),
\end{equation}
where $W_r$ and $U_r$ are the weight matrices, and $b_r$ is the bias vector.
The update gate is given by
\begin{equation} \label{eq.10}
z_t = \sigma(W_zx_t+U_zh_{t-1}+b_z),
\end{equation}
where $b_z$ is the bias, and $W_z$ and $U_z$ are the weight matrices.
The block output at time $t$ is then
\begin{equation} \label{eq.11}
h_t = z_t \odot h_{t-1} + (1-z_t) \odot \tanh(W_hx_t + U_h(r_t \odot h_{t-1}) + b_h),
\end{equation}
where $W_h$ and $U_h$ are the weight matrices, and $b_h$ is the bias vector.

\subsection{Proposed Models} \label{section.ProposedModels}

The first proposed model employs two stages, an Autoencoder and LSTM.
First, the data set is divided into two sets, positive labeled data (normal) and negative labeled data (abnormal) to train the Autoencoder.
The Autoencoder output is then fed into the LSTM network for anomaly detection and classification.

The Autoencoder has two networks (positive and negative) with three hidden layers (two encoder and one decoder layers).
For the positive Autoencoder network, the positive labeled data is selected and text preprocessing
such as tokenization is applied and the letters changed to lowercase.
Next, sentences are truncated or padded to 40 words (100 words for the IMDB data set) and sentences containing less than 5 words are deleted.
The word frequency is computed and the data set is shuffled.
Then an encoder layer with 400 neurons and L1 regularizer is used, followed by an encoder
layer with 200 neurons.
Next, a decoder layer with 200 neurons is used and finally an output layer with the same input size is applied.
This model is trained with just the positive labeled data without the labels.
The categorical cross entropy loss function with ADAM optimizer is used to train this model.
To prevent overfitting, dropout with probability 0.8 between each layer is employed and early stopping is used.
The batch size is 128 and the maximum number of training epochs is 100.

After training, the positive Autoencoder output (which is the same size as the input) is labeled as positive.
The negative Autoencoder network has the same architecture as the positive network,
the only difference is that negative labeled data is used.
After training, the output is labeled as negative.
Now the two outputs are concatenated to obtain a single labeled data set for anomaly detection and binary classification.
Duplicates are removed and Gaussian noise with zero mean and variance 0.1 is added to avoid overfitting \citep{NIPS2017_7096}.

The LSTM network is now used for anomaly detection and classification.
This network has a single hidden layer.
First, the concatenated data set is divided into testing and training sets
with 95\% for testing and 5\% for training, and these sets are shuffled.
The training set is then divided into two sets with $5\%$ for training and $95\%$ for training validation
(except for the IMDB and Openstack data sets with $85\%$ for training and $15\%$ for validation).
This is input to a Keras\footnote{\url{https://github.com/keras-team/keras}} embedding layer which converts each element to a vector.
A hidden LSTM layer of size 100 is used to classify the data into two labels using softmax activation in the final layer.
Categorical cross entropy is used as the loss function and the ADAM optimizer is applied.
To prevent overfitting, dropout is used in each LSTM layer with probability 0.8 and early stopping is applied.
10-fold cross validation is used in training with a maximum of 100 epochs and batch size 128.

This model is called Auto-LSTM.
Replacing the LSTM network in this model with the BLSTM and GRU networks gives the Auto-BLSTM and Auto-GRU models, respectively.
All three models have the same hyperparameters as described above.
The Auto-GRU architecture with two Autoencoder networks and a GRU network for anomaly detection and classification is shown in Figure~\ref{fig.5}.

\afterpage{
\begin{landscape}
\begin{figure}
\centering
\includegraphics[scale=0.55]{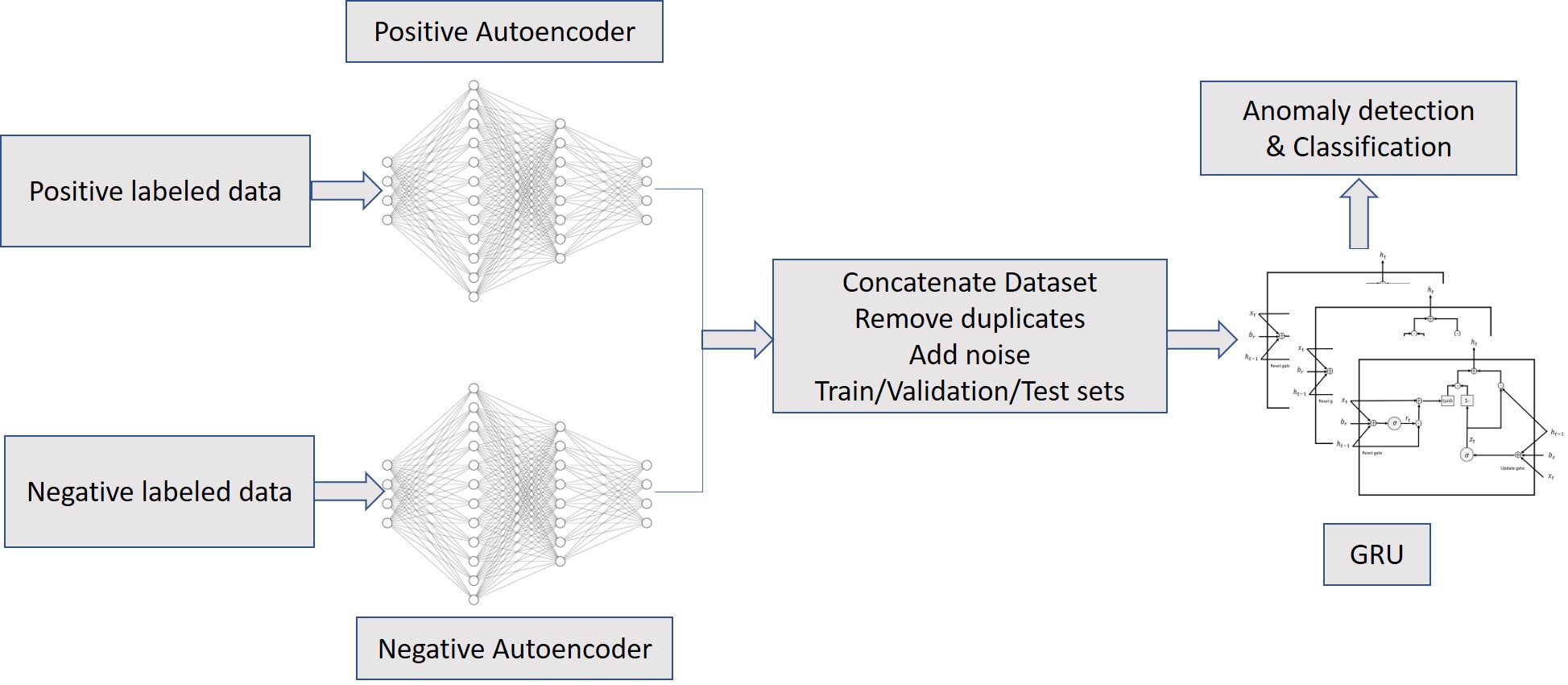}
\caption{The Auto-GRU architecture with two Autoencoder networks and a GRU network for anomaly detection and classification.} \label{fig.5}
\end{figure}
\end{landscape}}

\section{Results} \label{section.results}
In this section, the Auto-LSTM, Auto-BLSTM and Auto-GRU models are evaluated using four data sets, namely BGL, IMDB, Openstack and Thunderbird.
The following four criteria are used to evaluate the performance: accuracy, precision, recall and F-measure.
Accuracy is the fraction of input data that are correctly predicted and is given by
\begin{equation} \label{eq.12}
A=\frac{T_p+T_n}{T_p+T_n+F_p+F_n},
\end{equation}
where $T_p$ is the number of positive instances predicted by the model to be positive, $T_n$ is the number of negative instances predicted to be negative,
$F_p$ is number of negative instances predicted to be positive and $F_n$ is number of positive instances predicted to be negative.
Precision is given by
\begin{equation} \label{eq.13}
P=\frac{T_p}{T_p+F_p},
\end{equation}
and recall is expressed as
\begin{equation} \label{eq.14}
R=\frac{T_p}{T_p+F_n}.
\end{equation}
The F-measure is the harmonic mean of recall and precision
\begin{equation} \label{eq.15}
F=\frac{2 \times P \times R}{P+R}.
\end{equation}
All experiments were executed on the Compute Canada Graham cluster with 32 CPU cores, 124 GB memory and two P100 GPUs with Python in Keras.
The default hyperparameters were used for all data sets so they are not tuned for the proposed models.
Tables~\ref{table1} to \ref{table3} give the results for the BGL, IMDB, Openstack and Thunderbird data sets with the Auto-LSTM, Auto-BLSTM and Auto-GRU models, respectively.
For each data set, the average training accuracy, average validation accuracy, average training loss, testing accuracy, precision, recall and F-measure are shown.

\subsection{BGL} \label{section.BGL}

From the BlueGene/L (BGL) data set, 4,399,502 positive logs and 348,460 negative logs were obtained from the Autoencoder.
Of these, 11,869 logs were used for training, 225,529 for validation and the remaining 4,510,564 for testing.

With the Auto-LSTM model, the average training accuracy is 98.2\% and validation is 98.4\% with a standard deviation of 0.01 in 10-fold cross validation.
The average training loss is 0.09 with a standard deviation of 0.01.
The testing accuracy is 99.2\% with 98.0\% precision for the negative logs and 99.3\% for the positive logs, and
recall of 91.3\% and 99.8\% for negative and positive logs, respectively.
The F-measure is 94.5\% and 99.5\% for negative and positive logs, respectively.

With the Auto-BLSTM model, the average training accuracy is 98.4\% and validation is 98.7\% with a standard deviation of 0.01 in 10-fold cross validation.
The average training loss is 0.07 with a standard deviation of 0.01.
The testing accuracy is 99.3\% with precision of 98.9\% for negative logs and 99.4\% for positive logs, and recall of 92.1\% and
99.9\% for negative and positive logs, respectively.
The F-measure is 95.4\% and 99.6\% for negative and positive logs, respectively.

With the Auto-GRU model, the average training accuracy is 97.8\% and validation is 98.6\% with standard deviations of 0.02 and 0.01, respectively,
in 10-fold cross validation.
The average training loss is 0.07 with a standard deviation of 0.01.
The testing accuracy is 99.3\% with 98.9\% precision for negative logs and 99.3\% for positive logs, and
recall of 91.6\% and 99.9\% for negative and positive logs, respectively.
The F-measure is 95.1\% and 99.6\% for negative and positive logs, respectively.

The precision, recall and F-measure results for negative log messages are better than the
92\%, 91\% and 92\%, respectively, with the Improved K-Nearest Neighbor (IKNN) supervised algorithm \citep{S0218194020500114}.
Also, the precision, recall and F-measure results for negative logs are better than the
82.5\%, 94.7\% and 88.2\%, respectively, with the nLSALog algorithm \citep{8903291}, and the
83\%, 99\% and 91\%, respectively, with SVM unsupervised learning \citep{heExperienceReportSystem2016} (however, their recall is higher).

\subsection{IMDB} \label{section.IMDB}

The IMDB data set consists of 50,000 movie review sentences, with equal numbers that are positive and negative.
The Autoencoder reduced this to 49,565 sentences with 24,875 positive and 24,690 negative.
From these, 2,106 sentences were used for training, 372 for validation and the remaining 47,087 for testing.

With the Auto-LSTM model, the average training accuracy and average validation accuracy are 97.2\% and 97.9\%, respectively,
with standard deviations of 0.01 and 0.02, respectively, in 10-fold cross validation.
The average training loss is 0.08 with a standard deviation of 0.03.
The testing accuracy is 97.8\% with a precision of 97.9\% for negative labels and 97.7\% for positive labels, and
recall of 97.7\% and 97.9\% for negative and positive labels, respectively.
The F-measure is 97.8\% for both labels.

With the Auto-BLSTM model, the average training accuracy and average validation accuracy are 98.3\% and 99.1\%, respectively,
with a standard deviation of 0.01 in 10-fold cross validation.
The average training loss is 0.05 with standard deviation 0.01.
The testing accuracy is 98.8\% with a precision of 99.6\% for negative labels and 98.0\% for positive labels, and
recall of 98.0\% and 99.6\% for negative and positive labels, respectively.
The F-measure is 98.8\% and 98.8\% for the negative and positive labels, respectively.

With the Auto-GRU model, the average training accuracy and average validation accuracy is 97.5\% and 98.4\%, respectively, with a
standard deviation of 0.02 in 10-fold cross validation in the training.
The average training loss is 0.08 with a standard deviation of 0.04.
The testing accuracy is 98.8\% with a precision of 99.8\% for negative labels and 97.9\% for positive labels,
and recall of 97.8\% and 99.8\% for negative and positive labels, respectively.
The F-measure is 98.8\% and 98.8\% for the negative and positive labels, respectively.
The accuracy is better than the 93.2\% with a Convolutional Recurrent Network which is the combination of a Convolutional Neural Network and a Recurrent Neural Network \citep{hassanConvolutionalRecurrentDeep2018}.
The accuracy is also better than the 94.5\% with a Paragraph Vector and
89.1\% for a simple LSTM network \citep{hongAnalysisDeeplyLearned2015}.

\subsection{Openstack} \label{section.Openstack}

For the Openstack data set, 137,074 positive log messages and 18,434 negative log messages were obtained from the Autoencoder.
From these, 6,608 messages were used for training, 1,167 for validation and the remaining 147,733 logs for testing.

With the Auto-LSTM model, the average training accuracy is 98.5\% and average validation accuracy is 98.6\% with a
standard deviation of 0.01 in 10-fold cross validation.
The average training loss is 0.05 with a standard deviation 0.01.
The testing accuracy is 99.1\% with precision of 99.4\% for negative logs and 99.0\% for positive logs,
and recall of 92.8\% and 99.9\% for negative and positive logs, respectively.
The F-measure is 96.0\% and 99.5\% for negative and positive logs, respectively.

With the Auto-BLSTM model, the average training accuracy is 98.5\% and the average validation accuracy is 99.4\% with a
standard deviation of 0.01 in 10-fold cross validation.
The average training loss is 0.05 with a standard deviation of 0.02.
The testing accuracy is 99.4\% with precision of 99.6\% for negative logs and 99.3\% for positive logs,
and recall of 95.2\% and 99.9\% for negative and positive logs, respectively.
The F-measure is 97.3\% and 99.6\% for negative and positive logs, respectively.

With the Auto-GRU model, the average training accuracy is 98.4\% and the average validation accuracy is 97.2\% with
a standard deviation of 0.01 in 10-fold cross validation.
The average training loss is 0.05 with a standard deviation of 0.01.
The testing accuracy is 98.3\% with precision of 97.9\% for negative logs and 98.3\% for positive logs,
and recall of 87.1\% and 99.8\% for the negative and positive logs, respectively.
The F-measure is 92.2\% and 99.0\% for negative and positive logs, respectively.

The precision, recall and F-measure results for negative log messages are similar to the
94\%, 99\% and 97\% with the Deeplog network \citep{duDeepLogAnomalyDetection2017} (however, their recall is higher).

\subsection{Thunderbird} \label{section.Thunderbird}

For the Thunderbird data set, 3,000,000 positive log messages and 3,248,239 negative log messages were obtained from the Autoencoder output.
From these, 15,620 messages were used for training, 296,791 for validation and the remaining 5,935,828 for testing.

With the Auto-LSTM model, the average training accuracy is 97.3\% and the average validation accuracy is 98.9\% with a
standard deviation of 0.01 in 10-fold cross validation.
The average training loss is 0.09 with a standard deviation of 0.02.
The testing accuracy is 99.0\% with precision of 98.4\% and 99.8\% and recall of 99.8\% and 98.2\% for negative and positive logs, respectively.
The F-measure is 99.1\% and 99.0\% for negative and positive logs, respectively.

With the Auto-BLSTM model, the average training accuracy is 98.0\% and the
average validation accuracy is 99.6\% with a standard deviation of 0.01 in 10-fold cross validation.
The average training loss is 0.07 with a standard deviation of 0.04.
The testing accuracy is 99.4\% with precision of 99.0\% and 99.8\% and recall of 99.9\% and 98.9\% for negative and positive logs, respectively.
The F-measure is 99.4\% and 99.3\% for negative and positive logs, respectively.

With the Auto-GRU model, the average training accuracy is 97.5\% and the average validation accuracy is 99.3\% with a
standard deviation of 0.01 in 10-fold cross validation.
The average training loss is 0.08 with a standard deviation of 0.02.
The testing accuracy is 99.2\% with precision of 98.7\% and 99.9\% and recall of 99.9\% and 98.5\% for negative and positive logs, respectively.
The precision, recall and F-measure results for negative log messages are better than the
96\%, 96\% and 96\%, respectively, with the Improved K-Nearest Neighbor (IKNN) supervised algorithm \citep{S0218194020500114}.

\subsection{Discussion} \label{section.discussion}

The proposed models provided good results for all four data sets.
The results for the Auto-BLSTM model were slightly better than for the Auto-LSTM and Auto-GRU models except for the
IMDB data set which was the same as with Auto-GRU.
Further, Auto-GRU was slightly better than Auto-LSTM except for the Openstack data set.
In the proposed models, very little training data (less than 1\%) was used whereas deep learning algorithms typically need significant data for learning.
This was because the input data was features extracted so the LSTM (BLSTM and GRU) network needed to learn less data during training.
Although the focus here was on log anomaly detection and classification, the proposed models were evaluated using the IMDB data set to show that they
can also provide good results on other text classification tasks such as sentiment classification.
Hyperparameter tuning such as the learning rate, network structure and number of hidden layers should improve the results obtained.

Training the Autoencoder network is the most important stage.
With text data, words can be converted to digits using methods such as word frequency, Bag-of-Words or TF-IDF.
However, the relationship between the data with these methods is not the most suitable for the network.
In recent years, embedding methods such as Word2vec \citep{mikolovEfficientEstimationWord2013} and GloVe \citep{penningtonGloveGlobalVectors2014}
have been proposed to deal with the relationship between words.
In the proposed models, the original data is converted from text to digits using the word frequency.
Then the Autoencoder is trained and information extracted using this converted data as input which
provides a better relationship with the original data for machine learning algorithms.
This data is then used as input to the embedding method for algorithms such as LSTM, BLSTM and GRU networks.
These proposed models are particularly suited to log messages because these messages contain unstructured data
which can be challenging for anomaly detection and classification algorithms.

\section{Conclusions} \label{section.conclusion}

Anomaly detection and classification of log messages is a very important task in Machine Learning.
In this paper, the Auto-LSTM, Auto-BLSTM and Auto-GRU models were proposed for this purpose.
The first stage of these models employs an Autoencoder network to extract information and features from the input data
and the second stage is anomaly detection and classification with an LSTM, BLSTM or GRU network.
The proposed models were tested on four different data sets.
Three of these data sets are log messages for anomaly detection and classification while the fourth is a sentiment movie review classification data set.
The models were shown to provide good results for these data sets with only a very small portion used for training.
This indicates that these models can be used for tasks other than log message anomaly detection and classification such as text classification.

Future research can consider the effect of hyperparameter tuning on the proposed models.
Further, algorithms such as a Convolutional Neural Network (CNN) can be used for log message anomaly detection and classification.

%\begin{acknowledgements}
%\end{acknowledgements}

\section*{Conflict of interest}
The authors declare they have no conflict of interest.

\bibliographystyle{tfcad}      % basic style, author-year citations

\bibliography{xbibliography}   % name your BibTeX data base

\clearpage
\afterpage{
\begin{landscape}
\begingroup
\renewcommand{\arraystretch}{2.0}
\begin{table}[htbp]
\centering
\caption{Auto-LSTM results for the BGL, IMDB, Openstack and Thunderbird data sets (the numbers in parenthesis are standard deviation). Positive labels are denoted by 1 and negative labels by 0. }
\label{table1}
\begin{tabular}{ccccccccc}
\multicolumn{1}{c|}{\begin{tabular}[c]{@{}c@{}}\textbf{Data set}\\\textbf{ } \end{tabular}} & \multicolumn{1}{c|}{\begin{tabular}[c]{@{}c@{}}\textbf{Average}\\\textbf{ Training}\\\textbf{ Accuracy} \end{tabular}} & \multicolumn{1}{c|}{\begin{tabular}[c]{@{}c@{}}\textbf{Average}\\\textbf{ Validation}\\\textbf{ Accuracy} \end{tabular}} & \multicolumn{1}{c|}{\begin{tabular}[c]{@{}c@{}}\textbf{Average}\\\textbf{ Training}\\\textbf{ Loss} \end{tabular}} & \multicolumn{1}{c|}{\begin{tabular}[c]{@{}c@{}}\textbf{Testing}\\\textbf{ Accuracy} \end{tabular}} & \multicolumn{1}{c|}{\textbf{Label} } & \multicolumn{1}{c|}{\textbf{Precision} } & \multicolumn{1}{c|}{\textbf{Recall} } & \textbf{F-measure}   \\ 
\hline
\multirow{2}{*}{\textbf{BGL} }                                                              & \multirow{2}{*}{\begin{tabular}[c]{@{}c@{}}98.2\% \\ (0.01) \end{tabular}}                    & \multirow{2}{*}{\begin{tabular}[c]{@{}c@{}}98.4\%\\ (0.01) \end{tabular}}                       & \multirow{2}{*}{\begin{tabular}[c]{@{}c@{}}0.09 \\ (0.01) \end{tabular}}                  & \multirow{2}{*}{99.2\%}                                                                            & 0                                    & 98.0\%                                   & 91.3\%                                & 94.5\%               \\ 
\cline{6-9}
                                                                                            &                                                                                                                        &                                                                                                                          &                                                                                                                    &                                                                                                    & 1                                    & 99.3\%                                   & 99.8\%                                & 99.5\%               \\ 
\hline
\multirow{2}{*}{\textbf{IMDB} }                                                             & \multirow{2}{*}{\begin{tabular}[c]{@{}c@{}}97.2\%\\ (0.01) \end{tabular}}                     & \multirow{2}{*}{\begin{tabular}[c]{@{}c@{}}97.9\% \\ (0.02) \end{tabular}}                      & \multirow{2}{*}{\begin{tabular}[c]{@{}c@{}}0.08 \\ (0.03) \end{tabular}}                  & \multirow{2}{*}{97.8\%}                                                                            & 0                                    & 97.9\%                                   & 97.7\%                                & 97.8\%               \\ 
\cline{6-9}
                                                                                            &                                                                                                                        &                                                                                                                          &                                                                                                                    &                                                                                                    & 1                                    & 97.7\%                                   & 97.9\%                                & 97.8\%               \\ 
\hline
\multirow{2}{*}{\textbf{Openstack} }                                                        & \multirow{2}{*}{\begin{tabular}[c]{@{}c@{}}98.5\%\\ (0.01) \end{tabular}}                     & \multirow{2}{*}{\begin{tabular}[c]{@{}c@{}}98.6\% \\ (0.01) \end{tabular}}                      & \multirow{2}{*}{\begin{tabular}[c]{@{}c@{}}0.05\\ (0.01) \end{tabular}}                   & \multirow{2}{*}{99.1\%}                                                                            & 0                                    & 99.4\%                                   & 92.8\%                                & 96.0\%               \\ 
\cline{6-9}
                                                                                            &                                                                                                                        &                                                                                                                          &                                                                                                                    &                                                                                                    & 1                                    & 99.0\%                                   & 99.9\%                                & 99.5\%               \\ 
\hline
\multirow{2}{*}{\textbf{Thunderbird} }                                                      & \multirow{2}{*}{\begin{tabular}[c]{@{}c@{}}97.3\%\\ (0.01) \end{tabular}}                     & \multirow{2}{*}{\begin{tabular}[c]{@{}c@{}}98.9\%\\ (0.01) \end{tabular}}                       & \multirow{2}{*}{\begin{tabular}[c]{@{}c@{}}0.09 \\ (0.02) \end{tabular}}                  & \multirow{2}{*}{99.0\%}                                                                            & 0                                    & 98.4\%                                   & 99.8\%                                & 99.1\%               \\ 
\cline{6-9}
                                                                                            &                                                                                                                        &                                                                                                                          &                                                                                                                    &                                                                                                    & 1                                    & 99.8\%                                   & 98.2\%                                & 99.0\%              
\end{tabular}
\end{table}
\endgroup
\end{landscape}

\begin{landscape}
\begingroup
\renewcommand{\arraystretch}{2.0}
\begin{table}[htbp]
\centering
\caption{Auto-BLSTM results for the BGL, IMDB, Openstack and Thunderbird data sets (the numbers in parenthesis are standard deviation). Positive labels are denoted by 1 and negative labels by 0.}
\label{table2}
\begin{tabular}{ccccccccc}
\multicolumn{1}{c|}{\textbf{Data set} } & \multicolumn{1}{c|}{\begin{tabular}[c]{@{}c@{}}\textbf{Average}\\\textbf{ Training}\\\textbf{ Accuracy} \end{tabular}} & \multicolumn{1}{c|}{\begin{tabular}[c]{@{}c@{}}\textbf{Average}\\\textbf{ Validation}\\\textbf{ Accuracy} \end{tabular}} & \multicolumn{1}{c|}{\begin{tabular}[c]{@{}c@{}}\textbf{Average}\\\textbf{ Training}\\\textbf{ Loss} \end{tabular}} & \multicolumn{1}{c|}{\begin{tabular}[c]{@{}c@{}}\textbf{Testing}\\\textbf{ Accuracy} \end{tabular}} & \multicolumn{1}{c|}{\textbf{Label} } & \multicolumn{1}{c|}{\textbf{Precision} } & \multicolumn{1}{c|}{\textbf{Recall} } & \textbf{F-measure}   \\ 
\hline
\multirow{2}{*}{\textbf{BGL} }          & \multirow{2}{*}{\begin{tabular}[c]{@{}c@{}}98.4\% \\ (0.01) \end{tabular}}                    & \multirow{2}{*}{\begin{tabular}[c]{@{}c@{}}98.7\%\\ (0.01) \end{tabular}}                       & \multirow{2}{*}{\begin{tabular}[c]{@{}c@{}}0.07 \\ (0.01) \end{tabular}}                  & \multirow{2}{*}{99.3\%}                                                                            & 0                                    & 98.9\%                                   & 92.1\%                                & 95.4\%               \\ 
\cline{6-9}
                                        &                                                                                                                        &                                                                                                                          &                                                                                                                    &                                                                                                    & 1                                    & 99.4\%                                   & 99.9\%                                & 99.6\%               \\ 
\hline
\multirow{2}{*}{\textbf{IMDB} }         & \multirow{2}{*}{\begin{tabular}[c]{@{}c@{}}98.3\%\\ (0.01) \end{tabular}}                     & \multirow{2}{*}{\begin{tabular}[c]{@{}c@{}}99.1\%\\ (0.01) \end{tabular}}                       & \multirow{2}{*}{\begin{tabular}[c]{@{}c@{}}0.05 \\ (0.01) \end{tabular}}                  & \multirow{2}{*}{98.8\%}                                                                            & 0                                    & 99.6\%                                   & 98.0\%                                & 98.8\%               \\ 
\cline{6-9}
                                        &                                                                                                                        &                                                                                                                          &                                                                                                                    &                                                                                                    & 1                                    & 98.0\%                                   & 99.6\%                                & 98.8\%               \\ 
\hline
\multirow{2}{*}{\textbf{Openstack} }    & \multirow{2}{*}{\begin{tabular}[c]{@{}c@{}}98.5\%\\ (0.01) \end{tabular}}                     & \multirow{2}{*}{\begin{tabular}[c]{@{}c@{}}99.4\%\\ (0.01) \end{tabular}}                       & \multirow{2}{*}{\begin{tabular}[c]{@{}c@{}}0.05\%\\ (0.02) \end{tabular}}                 & \multirow{2}{*}{99.4\%}                                                                            & 0                                    & 99.6\%                                   & 95.2\%                                & 97.3\%               \\ 
\cline{6-9}
                                        &                                                                                                                        &                                                                                                                          &                                                                                                                    &                                                                                                    & 1                                    & 99.3\%                                   & 99.9\%                                & 99.6\%               \\ 
\hline
\multirow{2}{*}{\textbf{Thunderbird} }  & \multirow{2}{*}{\begin{tabular}[c]{@{}c@{}}98.0\%\\ (0.01) \end{tabular}}                     & \multirow{2}{*}{\begin{tabular}[c]{@{}c@{}}99.6\%\\ (0.01) \end{tabular}}                       & \multirow{2}{*}{\begin{tabular}[c]{@{}c@{}}0.07 \\ (0.04) \end{tabular}}                  & \multirow{2}{*}{99.4\%}                                                                            & 0                                    & 99.0\%                                   & 99.9\%                                & 99.4\%               \\ 
\cline{6-9}
                                        &                                                                                                                        &                                                                                                                          &                                                                                                                    &                                                                                                    & 1                                    & 99.8\%                                   & 98.9\%                                & 99.3\%              
\end{tabular}
\end{table}
\endgroup
\end{landscape}

\begin{landscape}
\begingroup
\renewcommand{\arraystretch}{2.0}
\begin{table}[htbp]
\centering
\caption{Auto-GRU results for the BGL, IMDB, Openstack and Thunderbird data sets (the numbers in parenthesis are standard deviation). Positive labels are denoted by 1 and negative labels by 0.}
\label{table3}
\begin{tabular}{ccccccccc}
\multicolumn{1}{c|}{\begin{tabular}[c]{@{}c@{}}\textbf{Data set}\\\textbf{ } \end{tabular}} & \multicolumn{1}{c|}{\begin{tabular}[c]{@{}c@{}}\textbf{Average}\\\textbf{ Training}\\\textbf{ Accuracy} \end{tabular}} & \multicolumn{1}{c|}{\begin{tabular}[c]{@{}c@{}}\textbf{Average}\\\textbf{ Validation}\\\textbf{ Accuracy} \end{tabular}} & \multicolumn{1}{c|}{\begin{tabular}[c]{@{}c@{}}\textbf{Average}\\\textbf{ Training}\\\textbf{ Loss} \end{tabular}} & \multicolumn{1}{c|}{\begin{tabular}[c]{@{}c@{}}\textbf{Testing}\\\textbf{ Accuracy} \end{tabular}} & \multicolumn{1}{c|}{\textbf{Label} } & \multicolumn{1}{c|}{\textbf{Precision} } & \multicolumn{1}{c|}{\textbf{Recall} } & \textbf{F-measure}   \\ 
\hline
\multirow{2}{*}{\textbf{BGL} }                                                              & \multirow{2}{*}{\begin{tabular}[c]{@{}c@{}}97.8\% \\ (0.02) \end{tabular}}                    & \multirow{2}{*}{\begin{tabular}[c]{@{}c@{}}98.6\%\\ (0.01) \end{tabular}}                       & \multirow{2}{*}{\begin{tabular}[c]{@{}c@{}}0.07 \\ (0.01) \end{tabular}}                  & \multirow{2}{*}{99.3\%}                                                                            & 0                                    & 98.9\%                                   & 91.6\%                                & 95.1\%               \\ 
\cline{6-9}
                                                                                            &                                                                                                                        &                                                                                                                          &                                                                                                                    &                                                                                                    & 1                                    & 99.3\%                                   & 99.9\%                                & 99.6\%               \\ 
\hline
\multirow{2}{*}{\textbf{IMDB} }                                                             & \multirow{2}{*}{\begin{tabular}[c]{@{}c@{}}97.5\%\\ (0.02) \end{tabular}}                     & \multirow{2}{*}{\begin{tabular}[c]{@{}c@{}}98.4\%\\ (0.02) \end{tabular}}                       & \multirow{2}{*}{\begin{tabular}[c]{@{}c@{}}0.08 \\ (0.04) \end{tabular}}                  & \multirow{2}{*}{98.8\%}                                                                            & 0                                    & 99.8\%                                   & 97.8\%                                & 98.8\%               \\ 
\cline{6-9}
                                                                                            &                                                                                                                        &                                                                                                                          &                                                                                                                    &                                                                                                    & 1                                    & 97.9\%                                   & 99.8\%                                & 98.8\%               \\ 
\hline
\multirow{2}{*}{\textbf{Openstack} }                                                        & \multirow{2}{*}{\begin{tabular}[c]{@{}c@{}}98.4\%\\ (0.01) \end{tabular}}                     & \multirow{2}{*}{\begin{tabular}[c]{@{}c@{}}97.2\%\\ (0.01)\end{tabular}}                        & \multirow{2}{*}{\begin{tabular}[c]{@{}c@{}}0.05\\ (0.01) \end{tabular}}                   & \multirow{2}{*}{98.3\%}                                                                            & 0                                    & 97.9\%                                   & 87.1\%                                & 92.2\%               \\ 
\cline{6-9}
                                                                                            &                                                                                                                        &                                                                                                                          &                                                                                                                    &                                                                                                    & 1                                    & 98.3\%                                   & 99.8\%                                & 99.0\%               \\ 
\hline
\multirow{2}{*}{\textbf{Thunderbird} }                                                      & \multirow{2}{*}{\begin{tabular}[c]{@{}c@{}}97.5\%\\ (0.01) \end{tabular}}                     & \multirow{2}{*}{\begin{tabular}[c]{@{}c@{}}99.3\%\\ (0.01) \end{tabular}}                       & \multirow{2}{*}{\begin{tabular}[c]{@{}c@{}}0.08 \\ (0.02) \end{tabular}}                  & \multirow{2}{*}{99.2\%}                                                                            & 0                                    & 98.7\%                                   & 99.9\%                                & 99.3\%               \\ 
\cline{6-9}
                                                                                            &                                                                                                                        &                                                                                                                          &                                                                                                                    &                                                                                                    & 1                                    & 99.9\%                                   & 98.5\%                                & 99.2\%              
\end{tabular}
\end{table}
\endgroup
\end{landscape}}

\end{document}